\documentclass[11pt]{article}
\usepackage[utf8]{inputenc}
\usepackage[round]{natbib}
\usepackage[margin=1in]{geometry}
\usepackage{color}
\usepackage{xcolor}
\PassOptionsToPackage{dvipsnames,svgnames,table}{xcolor}
\definecolor{darkblue}{rgb}{0,0,1}
\usepackage[utf8]{inputenc} % allow utf-8 input
\usepackage[T1]{fontenc}    % use 8-bit T1 fonts
\usepackage[colorlinks=true,
            urlcolor=purple,
            linkcolor=red,
            citecolor=blue]{hyperref}       % hyperlinks
\usepackage{url}            % simple URL typesetting
\usepackage{booktabs}       % professional-quality tables
\usepackage{amsfonts}       % blackboard math symbols
\usepackage{nicefrac}       % compact symbols for 1/2, etc.
\usepackage{microtype}      % microtypography
\usepackage{graphicx}
\usepackage{subcaption}
\usepackage{amsmath}
\usepackage{amssymb}
\usepackage{amsthm}
\usepackage{wrapfig}
\usepackage{algorithm}
\usepackage{algpseudocode}

\def\argmin{\mathop{\rm argmin}\nolimits}

\newcommand{\bx}{\boldsymbol{x}}
\newcommand{\by}{\boldsymbol{y}}

\newcommand{\bX}{\boldsymbol{X}}

\newcommand{\btheta}{\boldsymbol{\theta}}

\newcommand{\bTheta}{\boldsymbol{\Theta}}

\newcommand{\I}{\mathcal{I}}
\newcommand{\E}{\mathbb{E}}

\newcommand{\J}{\mathcal{J}}
\newcommand{\cL}{\mathcal{L}}
\newcommand{\cO}{\mathcal{O}}

\newcommand{\tm}{\widehat{\bTheta}_n^{(\text{MoM})}}
\usepackage{bm}
\usepackage{bbm}

\newcommand{\sP}{\mathbbm{P}}

\usepackage{booktabs}
\usepackage{makecell}
\usepackage{multirow}

\newcommand{\one}{\mathbbm{1}}
\usepackage{algorithm}
\usepackage{algpseudocode}

\makeatletter
\newenvironment{breakablealgorithm}
  {% \begin{breakablealgorithm}
   \begin{center}
     \refstepcounter{algorithm}% New algorithm
     \hrule height.8pt depth0pt \kern3pt% \@fs@pre for \@fs@ruled
     \renewcommand{\caption}[2][\relax]{% Make a new \caption
       {\raggedright\textbf{\fname@algorithm~\thealgorithm} ##2\par}%
       \ifx\relax##1\relax % #1 is \relax
         \addcontentsline{loa}{algorithm}{\protect\numberline{\thealgorithm}##2}%
       \else % #1 is not \relax
         \addcontentsline{loa}{algorithm}{\protect\numberline{\thealgorithm}##1}%
       \fi
       \kern2pt\hrule\kern2pt
     }
  }{% \end{breakablealgorithm}
     \kern8pt\hrule\relax% \@fs@post for \@fs@ruled
   \end{center}
  }
\makeatother

\usepackage{comment}

\makeatletter
\renewcommand{\fnum@algorithm}{\fname@algorithm}
\makeatother

\newtheorem{thm}{Theorem}[section]

\newtheorem{cor}{Corollary}[section]
\newtheorem{assumption}{A\hspace{-4pt}}
\newtheorem{remark}{\textbf{Remark}}

\usepackage{mathrsfs}

\allowdisplaybreaks

\title{\textbf{Dirichlet Process-based Robust Clustering using the Median-of-Means Estimator}}

\usepackage[max2]{authblk}
\author[1]{Supratik Basu$^*$}
\author[2]{Jyotishka Ray Choudhury$^*$}
\author[3]{Debolina Paul}
\author[4]{Swagatam Das}

\affil[1]{Department of Statistical Science, \textsc{Duke University}, NC, USA}
\affil[2]{School of Industrial and Systems Engineering, \textsc{Georgia Institute of Technology}, GA, USA}
\affil[3]{Machine Learning Research Laboratory, Electronics and Communication Sciences Unit,\newline\textsc{Indian Statistical Institute}, Kolkata, India}
\affil[4]{Electronics and Communication Sciences Unit, \textsc{Indian Statistical Institute}, Kolkata, India}

\date{\vspace{-5ex}}
\begin{document}

\maketitle

% remove hyperref warnings
\makeatletter\def\Hy@Warning#1{}\makeatother 

\def\thefootnote{*}
\footnotetext{Joint first authors contributed equally to this research.}
\def\thefootnote{\arabic{footnote}}

\begin{abstract}
Clustering stands as one of the most prominent challenges in unsupervised machine learning. Among centroid-based methods, the classic $k$-means algorithm, based on Lloyd's heuristic, is widely used. Nonetheless, it is a well-known fact that $k$-means and its variants face several challenges, including heavy reliance on initial cluster centroids, susceptibility to converging into local minima of the objective function, and sensitivity to outliers and noise in the data.
When data contains noise or outliers, the Median-of-Means (MoM) estimator offers a robust alternative for stabilizing centroid-based methods. On a different note, another limitation in many commonly used clustering methods is the need to specify the number of clusters beforehand. Model-based approaches, such as Bayesian nonparametric models, address this issue by incorporating infinite mixture models, which eliminate the requirement for predefined cluster counts.
Motivated by these facts, in this article, we propose an efficient and automatic clustering technique by integrating the strengths of model-based and centroid-based methodologies. Our method mitigates the effect of noise on the quality of clustering; while at the same time, estimates the number of clusters. Statistical guarantees on an upper bound of clustering error, and rigorous assessment through simulated and real datasets, suggest the advantages of our proposed method over existing state-of-the-art clustering algorithms.
\end{abstract}

\section{Introduction}
\label{sec:intro}

Within the fields of machine learning, data mining, and statistics, clustering stands out as a very prominent challenge in the domain of unsupervised learning. Its focus lies in employing methodologies to reveal underlying patterns, called clusters, within datasets. These clusters are defined so that data points grouped within the same cluster demonstrate a degree of internal similarity \citep{Xu2015}. Conversely, data points originating from separate clusters are expected to exhibit notable dissimilarity. Typically, data points are portrayed as vectors encompassing variables, referred to as features, in the machine learning community.
% Clustering is a fundamental problem in unsupervised learning and is ubiquitous in various applications and domains (Chandola et al., 2009), (Pediredla \& Seelamantula, 2011), (Jain, 2010), (Dhillon et al., 2003). 

The $k$-means algorithm \citep{llyod-kmeans} stands as a classic and extensively used clustering technique. When given a specific number of clusters, say $K$, the $k$-means algorithm iterates through two key steps: cluster assignment, where each data point is assigned to the cluster with the nearest centroid based on Euclidean or $\ell_2$ distance, and computing the cluster centroids, which involves placing each cluster's centroid at the sample mean of the points assigned to that cluster over the course of the current iteration. Given a dataset $\mathcal{X}=\left\{{\bX}_1, \ldots, {\bX}_n\right\}$, the $k$-means algorithm attempts to partition $\mathcal{X}$ into $K$ mutually exclusive classes by optimizing the objective function: 
\begin{equation}
    f_{\operatorname{KM}}(\bTheta) = \sum_{i=1}^n \displaystyle\min_{1 \leq j \leq K} ||\bX_i-\btheta_j||_2^2.
\end{equation}
where $\bTheta=\left\{\btheta_1, \btheta_2, \ldots, \btheta_K\right\}$ is the set of centroids corresponding to each of the $K$ clusters, and $\|\cdot\|_2$ is the usual $\ell_2$ norm. This optimization seeks to minimize the within-cluster variability.
Unfortunately, $k$-means and its variants suffer from several well-documented limitations, such as significant reliance on the initial selection of cluster centroids \citep{pmlr-v70-bachem17b}, tendency to converge to suboptimal local minima rather than the global minimum of the objective function \citep{xu-lange-2019}, and importantly, high sensitivity to outliers \citep{K-means-outliers}. Moreover, $k$-means performs poorly when the clusters are non-spherical \citep{spectral-andrew-ng}, and even when the clusters are spherical but with unequal cluster radii and densities \citep{Raykov2016-mg}. Apart from $k$-means, some popular clustering methods include its improved version $k$-means$++$ \citep{Arthur2007kmeansTA}, as well as $k$-medians \citep{bradley-k-median,k-median}, $k$-modes \citep{Chaturvedi2001}, $k$-Harmonic Means \citep{Zhang-1999-KHM}, etc.

%In particular, a drawback we seek to address in this article is the implicit assumption behind $k$-means that the data can be clustered spherically, which works well in Gaussian settings but can fail to separate even simple data examples otherwise (Ng et al., 2002).

Another major shortcoming of these algorithms used in practice is that most of them explicitly presuppose the number of clusters. The most commonly recognized algorithms such as $k$-means clustering, spectral clustering \citep{spectral-andrew-ng}, MinMax $k$-means clustering \citep{minmax-km}, Gaussian mixture models suffer from this issue. 

It is well-established in the machine learning community that Bayesian approaches generally offer room for more flexible models in various settings. For instance, the Dirichlet process mixture model \citep{hjort_holmes_müller_walker_2010}, which is notably a Bayesian nonparametric model, gives rise to infinite mixture models that do not require the number of clusters in the dataset to be supplied beforehand. \citep{DP-Means} considered such an approach that bridges the concepts of $k$-means and Gaussian mixture models \citep{bishop2006pattern,murphy2018machine}. Nevertheless, their method, called DP means, exhibits flexibility in guessing an optimal number of clusters, the algorithm utilizes the cluster average, i.e., arithmetic mean of the data points within the cluster, for centroid updation, compromising its performance specifically on noisy or outlier-laden datasets.

\begin{figure*}[t]
    \centering
    \includegraphics[width = \textwidth]{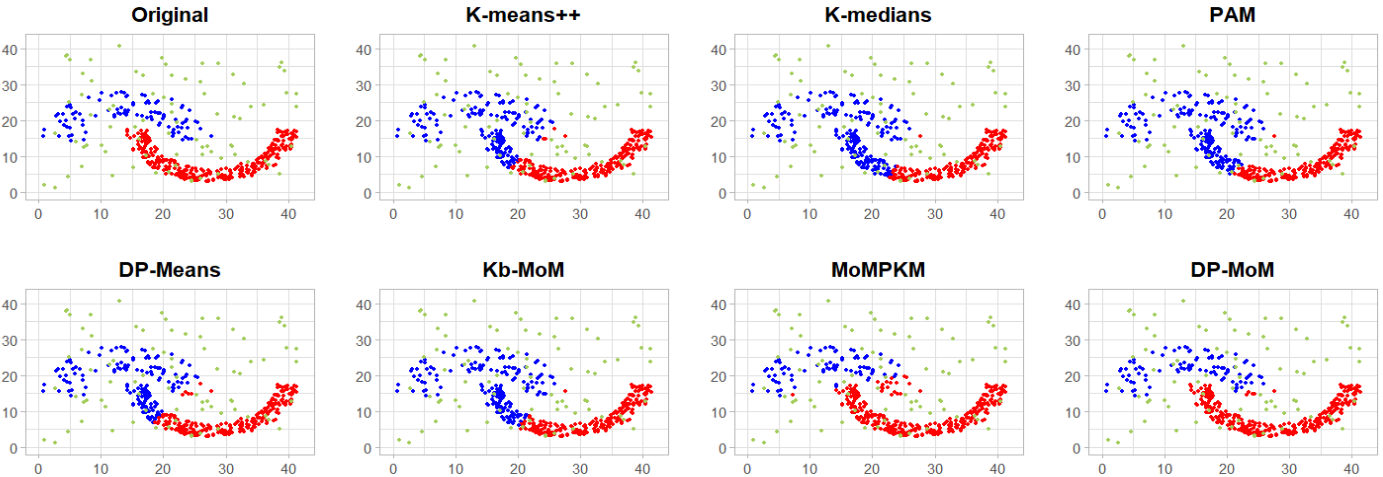}
    \caption{Several state-of-the-art clustering methods fail to achieve proper clustering in presence of noisy observations (light green in color), while the performance of DP-MoM, our proposed algorithm, is nearly optimal.}
    \label{fig:jain-out}
\end{figure*}

In this article, we address these challenges by fusing two prominent clustering methodologies: centroid-based and model-based. Our proposed algorithm, DP-MoM, is designed to excel in scenarios involving noisy or outlier-affected data, courtesy of its utilization of the Median-of-Means (MoM) estimator \citep{nemirovsky1983wiley,devroye-MoM}. Additionally, DP-MoM offers the advantage of not necessitating a predefined number of clusters. To demonstrate the efficacy of our approach, we present a compelling example. We introduce randomly generated noisy observations into the \textit{Jain} dataset \citep{jain-dataset} (refer to the Experiments section for detailed information) and subsequently apply various cutting-edge algorithms, including our proposed DP-MoM, to evaluate their respective performances on the original dataset. As illustrated in Figure \ref{fig:jain-out}, DP-MoM showcases notably superior clustering accuracy compared to existing algorithms.

\section{Background}
\label{gen_inst}

\subsection{Clustering based on Dirichlet Process}

\cite{DP-Means} introduced a method utilizing Gibbs sampling, serving as a Bayesian counterpart to representing $k$-means with a mixture of Gaussians. We assume the following model to capture the cluster structure of the dataset and whose limiting case reduces to the Dirichlet Process Mixture models:
\begin{align*}
\boldsymbol{\theta}_1, \ldots, \boldsymbol{\theta}_{{k}} \sim {G}_0 &, ~\pi \sim \operatorname{Dir}\left({k}, \pi_0\right), \\
{z}_1, \ldots, {z}_{{n}} \sim &\operatorname{Discrete}(\pi), \\
{\bX}_i \sim \mathcal{N}\left(\boldsymbol{\theta}_{{z}_{{i}}}, \sigma I\right) &\quad\forall i = 1,2,\ldots,n
\end{align*}
where $\boldsymbol{\theta}_{{j}}$ 's are the cluster centroids, ${G}_0$ is taken to be a $\mathcal{N}(0, \rho \mathbf{I})$ prior where $I$ denotes the identity matrix of appropriate order. $\operatorname{Dir}\left({k}, \pi_0\right)$ denotes the Dirichlet distribution, where $\pi$ is the mixture probability with $\pi_0=\frac{\alpha}{k} \mathbf{1}$. Here, $\mathbf{1}$  denotes the vector (of appropriate order) of all $1$'s. For $i = 1,2,\ldots,n$, ${z}_{{i}}$ denotes the label assigned to the data points ${\bX}_{{i}}$, and  $\operatorname{Discrete}(\pi)$ indicates that ${z}_{{i}}$ takes the value $j$ with probability $\pi_{{j}}$, for ${j}=1, \ldots, {k}$.

The hard clustering algorithm, called \textit{DP-means}, proposed by \cite{DP-Means} is essentially the case when $\sigma \to 0$. This limiting case boils down to minimizing the objective:
\begin{equation}\label{dp-means-obj}
    f_{\operatorname{DP}} (\bTheta, k)=\sum_{i=1}^n \min _{1 \leq j \leq k}\left\|{\bX}_i-\boldsymbol{\theta}_j\right\|_2^2+\lambda k.
\end{equation}

The minimization of the function in \eqref{dp-means-obj} with respect to $k$ is performed iteratively. At each step of the algorithm, the distance from each data point to its closest cluster centroid is determined. Subsequently, each point is assigned to the cluster with the closest centroid, unless this distance exceeds $\lambda$. In such an instance, a new cluster is initialized with the data point serving as the centroid of the newly created cluster. This algorithm determines the number of clusters in a dataset without necessitating prior knowledge of $k$. It maintains the simplicity inherent in Lloyd's approach, while ensuring effectiveness even in situations where the true number of clusters is unknown.

\subsection{Median-of-Means (MoM)}

Let us consider a simple scenario first where we observe $\bX_1,\bX_2,\ldots,\bX_n\sim F$ with a goal to estimate the mean of the distribution $F$, i.e., $\mu_0=\mathbb{E} [\bX_1]=\int x\,dF(x)$.

We employ the median-of-means (MoM) estimator to estimate $\mu_0$ as follows: Assume that the sample size $n=bL$ where $L$ is the number of buckets (disjoint subsamples) and $b$ is the size of each bucket. We first split the data randomly into $L$ partitions (or buckets), and calculate the mean of the data points belonging to each partition. This gives rise to estimators $\widehat{\mu}_1,\widehat{\mu}_2,\ldots,\widehat{\mu}_L$. The MoM estimator defined to be the median of these $b$ many mean estimators, namely,
\begin{equation}\label{MoM-defn}
\widehat{\mu}_{\text{MoM}}=\operatorname{median} (\widehat{\mu}_1,\widehat{\mu}_2,\ldots,\widehat{\mu}_L).
\end{equation}

As opposed to the sample mean which has a finite sample breakdown point of $0$, the MoM estimator is far more robust, exhibiting a breakdown point of $[(L-1)/2]/n$, thereby guaranteeing stability when the number of buckets is of the order of the number of data points i.e. $L=\mathcal{O}(n)$. Another reason why this estimator is of such interest (apart from being a robust estimator) is that given $\operatorname{var}(\bX_1)=\sigma^2<\infty$ in the finite sample case, it satisfies concentration inequalities that aid in proving consistency results thereby leading to stable and consistent MoM-based estimators.
% \citep{lerasle2019lecture}: 
% \begin{equation}
%     \mathbb{P}(|\widehat{\mu}_\text{MoM}-\mu_0| > \varepsilon) \leqslant e^{-2L} \left(\frac{1}{2}-\frac{L}{n}\frac{\sigma^2}{\varepsilon^2}\right)^2 \text{for all } n=bL.
% \end{equation}

In case of centroid-based clustering, the median-of-means estimator is used as follows: We first partition the dataset into $L$ subparts. In each iteration, we calculate the mean objective function value for each bucket $l$, $\frac{1}{b}\sum_{i \in B_{l}}f_{{\Theta}}(\bX_i)$ (where $\Theta$ is the collection of centroids from the previous iteration) and choose the bucket $L_t$ in such a way that
\begin{equation}\label{eq4}
    \frac{1}{b}\sum_{i \in B_{L_t}}f_{{\bTheta}}(\bX_i)=\underset{1\leqslant j\leqslant L}{\operatorname{median}}\left[\displaystyle \frac{1}{b}\sum_{i \in B_{j}}f_{{\bTheta}}(\bX_i)\right].
\end{equation}
The centroids $\Theta$ are recalculated based on the observations in the bucket $B_{L_t}$ and all the observations are clustered using these centroids.

\section{Dirichlet Process Clustering with MoM}

\subsection{Problem Formulation}
The problem posed to us is that of partitioning a given dataset $\mathcal{X} = \{\bX_1, \bX_2, \ldots, \bX_n\}\subset \mathbb{R}^p$ into natural, disjoint clusters such that the variance within each cluster is minimized at the same time maximizing the inter-partition variability.

In the context of centroid-based clustering, the $j^{th}$ cluster is represented by its centroid $\boldsymbol{\theta}_j$. The concept of "closeness" is quantified by utilization of a Bregman divergence \citep{BREGMAN1967200} as a dissimilarity measure. Let us denote the set of all non-negative real numbers by $\mathbb{R}^{+}_0$. Any function $\phi: \mathbb{R}^p \rightarrow \mathbb{R}$ that is convex and differentiable, gives rise to the Bregman divergence $d_\phi: \mathbb{R}^p \times \mathbb{R}^p \rightarrow \mathbb{R}^{+}_0$ defined as
\begin{equation}
    d_\phi(\boldsymbol{x}, \boldsymbol{y})=\phi(\boldsymbol{x})-\phi(\boldsymbol{y})-\langle\nabla \phi(\boldsymbol{y}), \boldsymbol{x}-\boldsymbol{y}\rangle.
\end{equation}

The Bregman divergence that we shall be using in our framework is the Euclidean distance, generated by $\phi(\boldsymbol{u})=\|\boldsymbol{u}\|_2^2$; although it can be generalized to any valid Bregman divergence. Prior knowledge of the number of centroids $k$ enables us to accomplish clustering by minimizing the objective function
\begin{equation}\label{ob}
    f_{\boldsymbol{\Theta}}(\boldsymbol{X}) := \frac{1}{n} \sum_{i=1}^n \Psi\left(d_\phi\left(\boldsymbol{X}_i, \boldsymbol{\theta}_1\right), d_\phi\left(\boldsymbol{X}_i, \boldsymbol{\theta}_2\right), \ldots, d_\phi\left(\boldsymbol{X}_i, \boldsymbol{\theta}_k\right)\right).
\end{equation}

Here, $\Psi: {\mathbb{R}^{+}_0}^k \rightarrow \mathbb{R}^{+}_0$ is the function $\min_{1\le j\le k} d_{\phi}(\bm{X},\bm{\theta}_j)$. In our case, we will seek to minimize the objective function
\begin{equation}\label{obj}
    h(\bTheta):=\operatorname{median}\left(\frac{1}{b} \sum_{i \in B_1} f_{\boldsymbol{\Theta}}\left(\boldsymbol{X}_i\right), \frac{1}{b} \sum_{i \in B_2} f_{\boldsymbol{\Theta}}\left(\boldsymbol{X}_i\right), \ldots, \frac{1}{b} \sum_{i \in B_L} f_{\boldsymbol{\Theta}}\left(\boldsymbol{X}_i\right)\right) + \lambda k
\end{equation}
with respect to both $\{\bm{\theta_j}\}_{1\le j\le k}$ and $k$.

\subsection{Optimization}

Optimizing the above objective is achieved using gradient-based methods. In our case, we employ the \textit{AdaGrad} algorithm \citep{duchi2011adaptive} for the said purpose. The centroids are updated as follows:
\begin{equation}
    \boldsymbol{\theta_j}^{(t+1)} := ~\boldsymbol{\theta}_j^{(t)} - \dfrac{\eta}{\sqrt{\varepsilon + \sum_{t' = 1}^t \left\|g_j^{(t')}\right\|^2}} \cdot g_j^{(t)},
\end{equation}
with 
\begin{equation}
    g_j^{(t)} = \frac{1}{b}\sum_{i \in B_{l_t}} 2(\boldsymbol{\theta}_j^{(t)} - \bX_i)\cdot \mathbf{I}_{\{\bX_i \in \mathcal{C}_j\}}
\end{equation}
where $\mathbf{I}_{\{\bX_i \in \mathcal{C}_j\}}=1$ if $\bX_i\in \mathcal{C}_j$, and $0$ otherwise.

% \begin{equation}
%     \mathbf{I}_{\{\bX_i \in \mathcal{C}_j\}}=\begin{cases}
% 1, & \text{if $u_{ij} = 1$}\\
% 0, & \text{otherwise.}
% \end{cases}
% \end{equation}

\subsection{Algorithm}

\begin{breakablealgorithm}%[!htb]
\caption{Dirichlet Process Clustering using Median-of-Means (DP-MoM)}
\label{algo1}
\flushleft{
\vspace{-2mm}\textbf{Input}: Data matrix $\mathcal{X}$, Penalty parameter $\lambda$, $\varepsilon$, Learning Rate $\eta$, Tolerance $\delta$.\\
\textbf{Output}: Number of clusters $k$, Cluster assignments $\mathcal{U}$, Cluster centroids $\bTheta$.\\
\textbf{Initialization}: Randomly divide $\{1,\dots,n\}$ into $L$ buckets of equal size. 
Set $\boldsymbol{\theta_1}=\frac{1}{n}\sum_{i = 1}^{n} \bX_i$, $k=1$, $b=\frac{n}{L}$, and $\mathcal{U}=\mathbf{1}$.\\
}
\vspace{2mm}
\begin{algorithmic}[1]
% \SetAlgoLined
 \While{$t<t_{\text{max}} ~\text{or}~ \left|\frac{h(\bTheta^{(t+1)})}{h(\bTheta^{(t)})} - 1\right| > \delta$}
  % instructions\;
  \For{every observation $\bX_i$}
  \State Compute $a_i$= $\min\left\{\|\bX_i - \boldsymbol{\theta_j}\|^2,~j = 1,\cdots,k\right\}$
  \If{$a_i > \lambda$}
  \State Set $k=k+1$, $\btheta_k=\bX_i$
  % \State Update $\mathcal{U}$ by $u_{ij} = \begin{cases} 1, \text{ if } j=k\\ 0, \text{ otherwise} \end{cases}$
  \State Update $\mathcal{U}$ by $u_{ij} = \mathbf{I}_{\{j=k\}}$
% \ENDIF
   \Else
   % \State Update $\mathcal{U}$ by $u_{ij}=\begin{cases} 1, \text{ if } j=\argmin_{1\leq c\leq k} ||\bX_i - \boldsymbol{\theta_c}||^2\\ 0, \text{ otherwise}\end{cases}$
   \State Update $\mathcal{U}$ by $u_{ij}=\mathbf{I}_{\{j=\argmin_{1\leq c\leq k} ||\bX_i - \boldsymbol{\theta_c}||^2\}}$
   \EndIf
   \EndFor
  \State Find $l_t \in \{1, 2, ... , L\}$ such that $\displaystyle\frac{1}{b}\sum_{i \in B_{l_t}} f_{\boldsymbol{\Theta^{(t)}}}(\bX_i) = \displaystyle\underset{1\leqslant j\leqslant L}{\operatorname{median}} \left[\displaystyle \frac{1}{b}\sum_{i \in B_{j}} f_{{\bTheta}}(\bX_i)\right]$
  \State For each $j\in \{1, 2, ..., k\}$: $\displaystyle g_j^{(t)} = \displaystyle\frac{1}{b}\sum_{i \in B_{l_t}} 2(\boldsymbol{\theta}_j^{(t)} - \bX_i)u_{ij}$
  \State Update ${\Theta}$ by $\displaystyle\boldsymbol{\theta_j}^{(t+1)} := ~\boldsymbol{\theta}_j^{(t)} - \dfrac{\eta}{\sqrt{\varepsilon + \sum_{t' = 1}^t ||g_j^{(t')}||^2}} \cdot g_j^{(t)}$
 \EndWhile
\end{algorithmic}
\end{breakablealgorithm}

% $$\text{THE ALGORITHM WILL BE STATED HERE}$$

Algorithm \ref{algo1} summarizes the pseudocode for the above procedure. The tuning parameter $\varepsilon$ is set to 1. The learning rate $\eta$ is typically chosen to be the power of 10 which is of the order of the squared maximum pairwise distance in the dataset, or one lower than that i.e. if the maximum squared separation between any two observations in the data is $D$, then we set $\eta = 10 ^{\lceil 2\log_{10} D \rceil /2}$ or $10^{\lceil 2\log_{10} D \rceil /2-1}$ depending on which of these values aids efficient clustering using our proposed method, where $\lceil \cdot \rceil$ represents the ceiling function. Our proposed framework enables us to automatically detect the appropriate number of clusters based on the value of the penalty parameter $\lambda$ which is optimized by grid-searching.

\subsection{Parameter Selection}
The first step in our proposed algorithm is partitioning the dataset randomly. This is achieved by choosing a permutation of the data points and then placing them in different buckets in the order of the permutation. Though this technique achieves randomness in terms of partitioning the data, arbitrary partitioning may lead to undesirable results, which is why the partitioning (or permutation for that matter) needs to be carefully chosen. It is pretty obvious that the buckets need to be fairly good representatives of the clusters for us to obtain accurate clustering. For this purpose, in each bucket of size $b$, we choose $b$ data points using a $k$-means++ type initialization to choose $b$ centers i.e., each data point has a probability proportional to its distance from the nearest centroid of being chosen. Once they are chosen, for the next bucket, we repeat this process leaving out the data points in all the previous buckets. This indexing scheme gives us a measure of control over the clustering and the subsequent grid-searching to determine the optimal value of the penalty parameter $\lambda$ and the number of buckets $L$.

We determine $\lambda_{\min}$ and $\lambda_{\max}$, the minimum and maximum pairwise squared distance between the data points, respectively. $11$ equally-spaced points $\lambda_{\min} =\lambda^1_1<\lambda^1_2< \cdots < \lambda^1_{10}<\lambda^1_{11}=\lambda_{\max}$ are picked, and the algorithm is run for these values and for all values of number of partitions, $L$, such that $2<L<\frac{n}{3}$. We select $\lambda^1_{i^*}$ corresponding to the most accurate clustering and divide its neighborhood $[\lambda^1_{i^*-1},\lambda_{i^*-1}]$  $\left([\lambda^1_1,\lambda^1_2]\operatorname{ or }[\lambda^1_{10}, \lambda^1_{11}]\right)$ into $20$ divisions and re-run the algorithm as we did in the interval $[\lambda_{\min}, \lambda_{\max}]$. We repeat this one more time, so that the feasible range for the penalty parameter $\lambda$ has been segmented to the order of $10^3$.

We then choose the $\lambda$ and $L$ values for which the best clustering accuracy is attained. We call them $\lambda_{opt}$ and $L_{opt}$. Since our proposed algorithm is a randomized one, we cannot readily conclude that $\lambda_{opt}$ and $L_{opt}$ are the only values corresponding to which we will obtain high clustering accuracy. In fact, for another repetition of the above experiment, we may not obtain an identical favorable permutation or the same optimal $\lambda$ and $L$ values. So, we repeat the aforementioned grid-searching experiment a number of times (say about 35 times) so that we may obtain a range of $\lambda$ and a range of $L$ that will be suitable to work with in order to derive the best results out of the proposed framework. We can however, restrict our search for the optimal value of $L$. Note that the breakdown point of the Median-of-Means estimator is $[(L-1)/2]/n$. In order for the centroid estimates to be stable and not break down in the presence of a little contamination, it is reasonable to assume that we should choose a value of $L$ such that $L=\mathcal{O}(n)$. We choose the median of the clustering accuracies, measured with the Adjusted Rand Index (ARI) \citep{Hubert1985} so obtained, as a representative of the clustering accuracy of the algorithm. When the ground truth is unavailable, we may use the t-SNE \citep{JMLR:v9:vandermaaten08a} plot of the data to form an idea of the ground truth. 

\subsection{Computational Complexity}
% \textcolor{red}{$\mathcal{O}(nKp)$ per iteration}
In each iteration, our algorithm first ascertains whether an increase in the number of clusters is needed. The centroids are recalculated thereafter, and the cluster assignments are made accordingly. This phase typically takes $\mathcal{O}(nCp)$ time steps to complete, where $C$ represents the number of clusters in that iteration. The calculations presented in the following section assume that the number of clusters is upper bounded by some finite $K<n$. Consequently, the worst case runtime of the DP-MoM algorithm remains $\mathcal{O}(nKp)$ for every iteration.

The computational complexity of the DP-means algorithm is comparable to that of DP-MoM, as each iteration requires $\mathcal{O}(nCp)$ steps to complete, with $C$ denoting the number of clusters in that specific iteration. On the contrary, $k$-means demands $\mathcal{O}(nkp)$ steps per iteration, with $k$ representing the predefined cluster count. This is typically slated to be lower than that of DP-means or DP-MoM. In the case where we set $k=K$ however, $k$-means will perform no more efficiently than DP-MoM in terms of runtime.

%%%%

\section{Theoretical Analysis}\label{sec:theory-new}
Let $\mathcal{M}$ denote the set of probability measures $P$ on $\mathbb{R}^p$ with bounded $\ell_2$ norm, i.e., for any random vector $\bX\sim P$, we have $\E \|\bX\|^2\le \gamma^2<\infty$ where $\|\cdot\|$ denotes the $\ell_2$ norm. We shall make the standard assumption that all the data points are independent and identically distributed (\textit{i.i.d.}) with finite squared $\mathcal{L}_2$ norm.

\begin{assumption}\label{ass-1-iid}
    $\bm{X}_1,\bm{X}_2,\ldots,\bm{X}_n \overset{\text{iid}}{\sim}P$ such that $P\in \mathcal{M}$.
\end{assumption}

We denote the empirical distribution derived from $\bX_1,\bX_2,\ldots,\bX_n$, by $P_n$, that is, $P_n(A) = \frac{1}{n}\sum_{i=1}^n \mathbf{I}\{\bX_i \in A\}$ for any Borel set $A$. 
For the sake of notational simplicity, we write $\mu f := \int f \, d\mu$. The quantities $P_n f_{\bTheta}$ and $P f_{\bTheta}$ are defined as follows: \[P_n f_{\bTheta}=\frac{1}{n}\sum_{i=1}^n f_{\bTheta}(\bX_i)~~\text{and}~~Pf_{\bTheta}=\E_{P} [f_{\bTheta}(\bX)].\]
Further let $\bTheta^{*}$ be the global minimizer of $P f_{\bTheta}$, and $\widehat{\bTheta}_n^{(MoM)}$ be the minimizer of \eqref{eq4}.

\begin{assumption}\label{ass-2-cluster-count}
    The number of clusters $k$ is bounded above by some finite $K\in \mathbb{N}$, where $K < n$.
\end{assumption}

The inherent dependency of the number of centroids on the cluster penalty parameter $\bm{\lambda}$ makes it possible for us to choose $\bm{\lambda}$ appropriately so that the cluster count doesn't exceed $K$. Moreover, we deduce from A\ref{ass-2-cluster-count} that, at a certain juncture, the number of centroids reaches a state of stability. In this state, it is only the cluster centroids themselves that undergo updates during each iteration, while the number of centroids remains constant. This will make our analysis independent of the penalty parameter $\bm{\lambda}$ as the term $k\bm{\lambda}$ is not subject to change after a finite number of iterations. Consequently, beyond a finite number of steps, the objective function effectively reduces to 
\begin{equation}\label{DP-MoM-obj}
    \operatorname{MoM}_L^n(\bTheta) := \operatorname{median} \left(\frac{1}{b}\sum_{i\in B_{1}}f_{\bm{\Theta}}(\bm{X}_i), \frac{1}{b}\sum_{i\in B_{2}}f_{\bm{\Theta}}(\bm{X}_i), \ldots, \frac{1}{b}\sum_{i\in B_{\ell}}f_{\bm{\Theta}}(\bm{X}_i)\right).
\end{equation} 

\begin{remark}
    In our framework, $\displaystyle |\Psi(\bm{x})-\Psi(\bm{y})|\le \|\bm{x}-\bm{y}\|_1$.
\end{remark}

To see this, recall the definition of $\Psi$ in \eqref{obj}, and note that $|\Psi(\bm{x})-\Psi(\bm{y})|=|\min_{1\le j\le K}x_j-\min_{1\le j\le K}y_j|$. Let us assume, without loss of generality, that $\min_{1\le j\le K}x_j\ge \min_{1\le j\le K}y_j$. Hence, $\min_{1\le j\le K}x_j-\min_{1\le j\le K}y_j\le x_{j^*}-y_{j^*}\le \|\bm{x}-\bm{y}\|_1$ where $j^* = \operatorname{argmin}_{1\le j\le K}y_j$ and the remark is seen to hold.

Since we have chosen an infinite support for $P$, following \cite{klochkov2020robustkmeansclusteringdistributions}, we can say that for $\bTheta^*$, there must exist a positive real number $M=M(p,k)$ such that $\|\btheta_j\|_2 \le M$ for every $\btheta_j\in\bTheta^*$ where $j=1,2,\ldots,k$. 

\subsection{Analysis under the Median-of-Means (MoM) Paradigm}

We represent the set of all inliers as $\{\bX_i\}_{i \in \I}$ and the outliers as $\{\bX_i\}_{i \in \cO}$. We make the following assumptions for determining the rate at which $|Pf_{\tm} - Pf_{\bTheta^\ast}|$ approaches $0$.

\begin{assumption}\label{ass-4-iid}
    $\{\bX_i\}_{i \in \I}\sim P$ are i.i.d. with $P \in \mathcal{M}$.
\end{assumption}

\begin{assumption}\label{ass-5-L}
    $\exists \, \eta > 0$ such that $L>(2+\eta)|\cO|$.
\end{assumption}

Assumption A\ref{ass-4-iid} ensures that the inliers arise independently from some distribution $P$. A\ref{ass-5-L} guarantees that at least half of the $L$ partitions are devoid of outliers. Such an assumption involving an upper bound on the number of outliers is essential as a high degree of contamination would imply that these so called `outliers' need to be treated as the data. This is, in fact, a milder requirement compared to the condition $L > 4|\cO|$ imposed in the recent work \citep{lecue2020robust}. Crucially, we highlight that no distributional assumptions are imposed on the outliers, permitting them to be unbounded, originate from heavy-tailed distributions, or exhibit any dependence structure among themselves. 

Since $\|\btheta\|_2\le M$ for all $\btheta\in \bTheta^*$, the search space for $\tm$ may be constrained to $\mathscr{G}$. Subsequently, we establish a high probability bound on $\sup_{\bTheta \in \mathscr{G}} |\text{MoM}^n_L (f_{\bTheta}) - Pf_{\bTheta} |$, and further use it to bound $|P f_{\tm} - P f_{\bTheta^\ast}|$. 
We %define $\delta:=2/(4+\eta) - |\cO|/L$, and 
use ``$\lesssim$'' to denote the fact that a quantity is lesser than a constant multiple of the other. % $\delta:= \frac{2}{4+\eta} - \frac{|\cO|}{L}$; 

\begin{thm}\label{thm-4-MoM}
Under A\ref{ass-4-iid}-A\ref{ass-5-L}, with probability at least $1-2e^{-2L \delta^2}$, 
\[ \sup_{\bTheta \in \mathscr{G}} \left|\text{MoM}^n_L (f_{\bTheta}) - Pf_{\bTheta} \right| \lesssim {M(M+2\gamma)}n^{-1/2}.\]
\end{thm}

\noindent
We present below a corollary that aids us in controlling the absolute difference $|P f_{\tm} - P f_{\bTheta^\ast}|$.

\begin{cor}\label{cor-mom-1}
    Under A\ref{ass-4-iid}-A\ref{ass-5-L}, with probability at least $1-2e^{-2L \delta^2}$, \[\left|P f_{\widehat{\bTheta}_n^{MoM}}-P f_{\bTheta^*}\right|\lesssim {M(M+2\gamma)}n^{-1/2}.\]
\end{cor}

\subsection{Asymptotic Properties: Consistency and Rate of Convergence}

We now consider the classical setting where $p$ is held constant, and demonstrate that the previously presented results imply consistency, with rate of convergence of the order of $\mathcal{O}(n^{-1/2})$. We first follow the same idea of convergence of $\widehat{\bTheta}_n^{\text{MoM}}$ to $\bTheta^\ast$ that is outlined in \cite{pollard1981strong}. Since the centroids are unique up to rearrangement of labels, our concept of dissimilarity
\[\mathcal{D}(\bTheta_1,\bTheta_2) = \min_{H \in \mathscr{P}_K} \|\bTheta_1 - H \bTheta_2\|_F \]
is considered over $\mathscr{P}_K$ the set of all real permutation matrices of order $K$, where $\|\cdot\|_F$ represents the
Frobenius norm. The sequence $\bTheta_n\to \bTheta$ if $\lim_{n\to \infty} \mathcal{D}(\bTheta_n,\bTheta)=0$. Following \cite{terada2014strong,chakraborty2020entropy}, we assume the identifiablity condition:

\begin{assumption}\label{ass-3-diss}
    $\forall$ $ \eta>0$, $\exists$ $\varepsilon>0$, such that $\, \mathcal{D}(\bTheta,\bTheta^\ast)> \eta$ $\implies$ $P f_{\bTheta} > P f_{\bTheta^\ast} + \varepsilon$ .
\end{assumption}

We now examine the consistency of $\widehat{\bTheta}_n^{\text{MoM}}$. In addition, we analyze the rate at which $|P f_{\widehat{\bTheta}_n^{\text{MoM}}} - P f_{\bTheta^\ast}|$ approaches $0$. Theorem \ref{cor-mom-2} affirms that strong consistency holds, with $\mathcal{O}(n^{-1/2})$ rate of convergence. Before delving further, recall that $X_n = \mathcal{O}_P(a_n)$ if the sequence $X_n/a_n$ is bounded in probability.

\begin{assumption}\label{ass-6-buckets}
    The number of buckets $L\to \infty$ as $n\to \infty$.
\end{assumption}

\begin{thm}\label{cor-mom-2}
    Under A\ref{ass-1-iid}-A\ref{ass-3-diss}, \[\left|P f_{\widehat{\bTheta}_n^{\operatorname{MoM}}}-P f_{\bTheta^*}\right|=\mathcal{O}_P(n^{-1/2})\] and consequently $P f_{\widehat{\bTheta}_n^{\operatorname{MoM}}}\overset{p}{\longrightarrow}P f_{\bTheta^*}$. Furthermore, under A\ref{ass-6-buckets}, we have $\widehat{\bTheta}_n^{\operatorname{MoM}}\overset{p}{\longrightarrow}\bTheta^*$.
\end{thm}

\section{Experiments}
\label{sec:experiments}
We now empirically compare our proposed framework with existing clustering approaches to thoroughly validate and evaluate its effectiveness. The accuracy of cluster assignments has been rigorously assessed using the Adjusted Rand Index (ARI), a robust measure of clustering performance.
Our evaluation encompasses an extensive array of competing centroid-based clustering methods, including renowned techniques such as $k$-means$++$, Sparse $k$-means (SKM) \citep{SKM-paper}, $k$-medians, Partition around Medoids (PAM) \citep{PAM-paper}, Robust Continuous Clustering (RCC) \citep{Shah2017-jj}, DP-means \citep{DP-Means}, $k$-bootstrap Median-of-Means ($K$-bMOM) \citep{brunetsaumard2020kbmom}, Median-of-Means with Power $k$-means (MoMPKM), and Ordered Weighted $l_1$ $k$-means (OWL $k$-means) \citep{pmlr-v206-chakraborty23a}. These state-of-the-art algorithms are benchmarked against the proposed DP-MoM algorithm across various experimental scenarios. The simulation experiments were conducted using a computer equipped with Intel(R) Core(TM)i3-7020U  2.30GHz  processor, 4GB RAM, 64-bit Windows 10 operating  system in the \texttt{R} programming language \citep{R-lang}. The relevant codes can be accessed at \url{https://github.com/jyotishkarc/Clustering-Techniques/tree/main/DP-MoM}.

Our experiments involve implementing the aforementioned techniques on a simulated dataset as well as several datasets from the UCI Machine Learning Repository\footnote{\url{https://archive.ics.uci.edu/}} and the Compcancer database\footnote{\url{https://schlieplab.org/Static/Supplements/CompCancer/}}. Owing to the fact that our clustering technique relies on randomization while partitioning the data into buckets, the accuracy measure has been computed as the median value of the obtained ARI over $30$ test runs. It was observed, for most of the datasets, that DP-MoM performed considerably better than its competitors in terms of ARI. Apart from this, two other experiments were conducted to assess the strength of the algorithm in terms of robustness and ability to detect clusters of various shapes that were in proximity to one another in terms of their pairwise Euclidean distances.

\subsection{Simulation Studies}
% \paragraph{Introducing Outliers in Real Data} \newline

\paragraph{Study of Robustness on Simulated Data:} $30$ data points are generated from each of the $4$ quadrants in the $2$-dimensional Euclidean plane using a special generation scheme. For the first quadrant, we generate $R_i \sim U(0, 1)$ and $\theta_i \sim U\left(\frac{\pi}{36}, \frac{17\pi}{36}\right)$. Once this is done for all $i=1,2,\ldots,30$, we set $X_i = (R_i\cos\theta_i, R_i\sin \theta_i)$ for all $i=1,2,\ldots,30$ as our data points in the first quadrant. In the other quadrants, we draw $R_i$ in the same way and generate $\theta_i$ uniformly from $\left(\frac{(j - 1)\pi}{2} + \frac{\pi}{36}, \frac{j\pi}{2} - \frac{\pi}{36}\right)$ for the $j^{th}$ quadrant. We place the data points lying in the same quadrant in the same cluster. Just like in the experiment using the \textit{Jain} dataset, outliers have been generated uniformly on $[-1, 1] \times [-1, 1]$. $15, 15,$ and $20$ outliers were introduced in three stages, respectively, so that the total number of data points stood at $135$, $150$, and $170$ respectively. While the efficiency of the other competing algorithms plummeted or showed erratic behavior (often combined with low clustering accuracy), the ARI corresponding to DP-MoM did not waver. Figure 2 depicts the superiority of our proposed algorithm over the other existing clustering techniques.

\begin{figure}[!htb]
    \centering
    \includegraphics[width=0.5\textwidth]{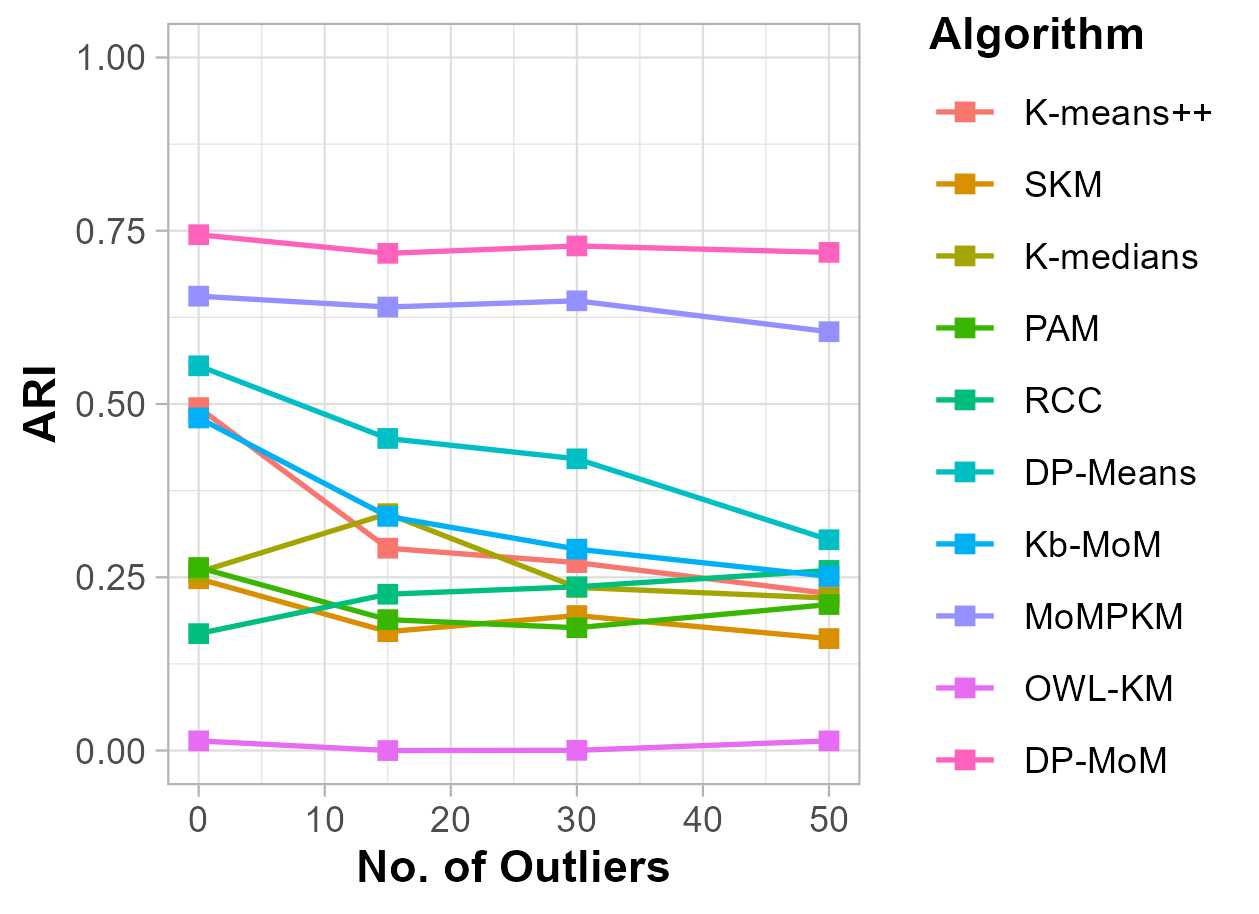}
    \caption{Line plots of ARI produced by different algorithms on simulated datasets, for increasingly higher number of outliers. DP-MoM is observed to perform uniformly better than all the competing methods.}
    \label{fig:plot-sim-ari}
\end{figure}

\subsection{Real Data Experiments}

\subsubsection{Introducing Outliers - A Case Study}

We have picked the dataset \textit{Jain} \citep{jain-dataset} for this experiment. \textit{Jain} is a $2$-dimensional dataset with $373$ data points. The $2$ natural clusters are shaped like boomerangs, as can be seen in Figure 1 in Section \ref{sec:intro}. The performance of the algorithms was assessed on the original dataset. Afterward, several outliers were uniformly generated throughout the range of the data. $20$ fresh outliers were introduced in each of $4$ stages and at each stage, the algorithms were pitted against each other again. Even with the introduction of $80$ outliers, DP-MoM remained remarkably robust, consistently achieving a clustering accuracy of nearly $0.9$ in terms of ARI  (while the maximum ARI achieved was above that figure in all but one stage). Conversely, many other competing algorithms struggled to maintain their performance in the face of increasing outlier counts. They exhibited significant fluctuations in ARI as the number of outliers rose. Even the ones that maintained stability could only muster a measly ARI of $0.42$, as did all the other competing techniques.

\begin{figure}[!htb]
    \centering
    \includegraphics[width=0.45\textwidth]{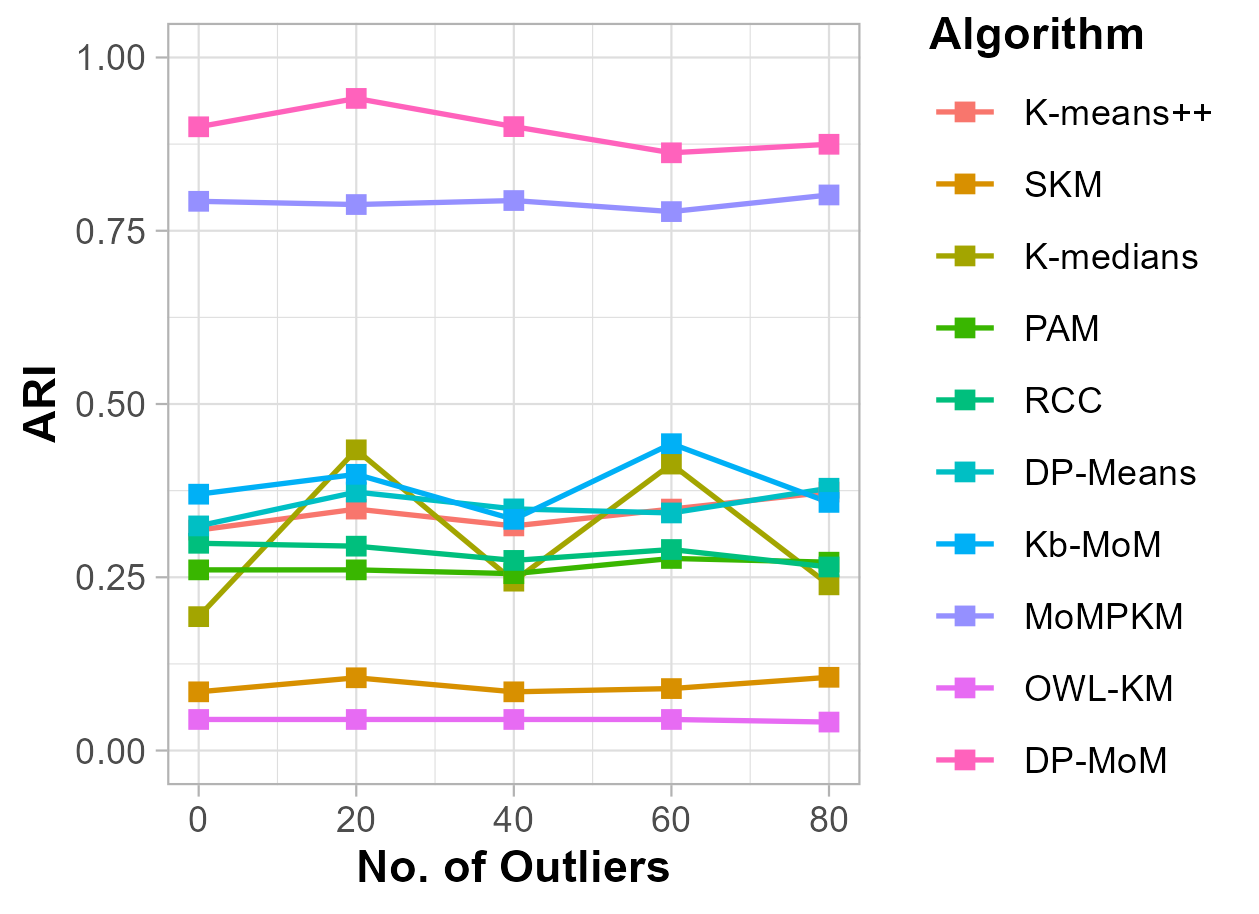}
    \caption{Line plots of ARI produced by different algorithms, for increasingly higher number of noisy observations introduced in the \textit{Jain} dataset. DP-MoM performs better than all the competing methods.}
    \label{fig:plot-jain-ari}
\end{figure}

\subsubsection{Further Experiments}

For a comprehensive performance evaluation of our proposed clustering algorithm in situations where the underlying data distributions are unknown, we implement DP-MoM on several real datasets from the Compcancer database and the UCI Repository. Additionally, we implement some state-of-the-art clustering algorithms mentioned at the start of this section, on the same datasets, and compare the corresponding ARI values against that of DP-MoM. Since DP-MoM is a randomized algorithm in the sense that its cluster assignment is dependent on the initial dataset partitioning into buckets, we implement DP-MoM on each dataset $30$ times independently, and report the median ARI. The same procedure is followed while reporting the ARI values for the competing algorithms.

\begin{table*}[!htb]
\renewcommand\theadfont{\bfseries}
\centering
\caption{ARI values corresponding to clustering via state-of-the-art algorithms as well as DP-MoM on UCI datasets}
% \vskip 1em
\resizebox{\linewidth}{!}{
\begin{tabular}{lccccccccccccc}
\toprule
\multirow[b]{2}{*}{\thead{\textbf{Dataset}}}
    &   \multicolumn{3}{c}{\thead{\textbf{Description}}}
        & \multicolumn{9}{c}{\thead{\textbf{State-of-the-Art Algorithms}}}
             & \multirow[b]{2}{*}{\thead{\textbf{DP-MoM}}}\\
\cmidrule(lr){2-4}\cmidrule(lr){5-13} %\cmidrule(lr){11-13}
% \toprule
% & $\mathbf{n}$ & $\mathbf{d}$ & $\mathbf{K}$ & \thead{KM++} & \thead{DPM} & \thead{KMed} & \thead{PAM} & \thead{K-bMoM} & \\
& \textit{\textbf{n}} & \textit{\textbf{p}} & \textit{\textbf{K}} & \thead{KM++} & \thead{SKM} & \thead{KMed} & \thead{PAM} & \thead{RCC} & \thead{DPM} & \thead{KbMoM} & \thead{MoMPKM} & \thead{OWL-KM}  & %\thead{DP-MoM}
\\
% \addlinespace
\midrule
\addlinespace
Iris           & 150  & 4  & 3 & 0.7237 & 0.7960 & 0.7515 & 0.6325 & 0.8090 & 0.7515          & 0.7565 & 0.8647          & 0.6339  & \textbf{0.9799}  \\
Glass          & 214 & 9  & 7 & 0.3728 & 0.3595 & 0.3367 & 0.3501 & 0.3930 & \textbf{0.4472} & 0.3467 & 0.2484          & 0.2659 & 0.3190          \\
WDBC           & 569 & 30 & 2 & 0.4223 & 0.4223 & 0.4603 & 0.4587 & 0.4146 & 0.4479          & 0.4560 & \textbf{0.6839} & 0.5897 & 0.6798           \\
E.Coli         & 336 & 7  & 8 & 0.5001 & 0.4918 & 0.5346 & 0.5407 & 0.5350 & 0.6663          & 0.6216 & 0.4952          & 0.2174 & \textbf{0.7835} \\
Wine           & 178 & 13 & 3 & 0.4140 & 0.4287 & 0.4226 & 0.4189 & 0.3564 & 0.4094          & 0.4227 & 0.5518          & 0.3470 & \textbf{0.5820} \\
Thyroid        & 215 & 5  & 3 & 0.3936 & 0.2145 & 0.1450 & 0.2144 & 0.5186 & 0.4971          & 0.3032 & 0.5995          & 0.4392  & \textbf{0.8842} \\
Zoo            & 101 & 16 & 7 & 0.7376 & 0.7516 & 0.6730 & 0.6566 & 0.7173 & 0.8270          & 0.4978 & 0.7603          & 0.8408 & \textbf{0.8477} \\
soybean & 47 & 35 & 4 & 0.7143 & 0.7138 & 0.7108 & 0.7437 & 0.8268 & 0.7368          & 0.82678 & 0.7417          & 0.5452 & \textbf{0.9533} \\
\addlinespace
\midrule
Average Rank        &          &                    &                   & 6.5625          & 6.1875  & 6.9375 & 6.5000          & 5.1875    & 4.6875            & 5.4375         & 4.2500          &  7.2500           & \textbf{2.0000}\\
\bottomrule
\end{tabular}}
\label{table:uci}
\end{table*}

\begin{table*}[!htb]
\renewcommand\theadfont{\bfseries}
\centering
\caption{ARI values corresponding to clustering via state-of-the-art algorithms as well as DP-MoM on Compcancer datasets}
% \vskip 1em
\resizebox{\linewidth}{!}{
\begin{tabular}{lccccccccccccc}
\toprule
\multirow[b]{2}{*}{\thead{\textbf{Dataset}}}
    &   \multicolumn{3}{c}{\thead{\textbf{Description}}}
        & \multicolumn{9}{c}{\thead{\textbf{State-of-the-Art Algorithms}}}
             & \multirow[b]{2}{*}{\thead{\textbf{DP-MoM}}}\\
\cmidrule(lr){2-4}\cmidrule(lr){5-13} %\cmidrule(lr){11-13}
% \toprule
% & $\mathbf{n}$ & $\mathbf{d}$ & $\mathbf{K}$ & \thead{KM++} & \thead{DPM} & \thead{KMed} & \thead{PAM} & \thead{K-bMoM} & \\
& \textit{\textbf{n}} & \textit{\textbf{p}} & \textit{\textbf{K}} & \thead{KM++} & \thead{SKM} & \thead{KMed} & \thead{PAM} & \thead{RCC} & \thead{DPM} & \thead{KbMoM} & \thead{MoMPKM} & \thead{OWL-KM}  & %\thead{DP-MoM}
\\
% \addlinespace
\midrule
\addlinespace
golub\_1999\_v2     & 72         & 1868                  & 3                  & 0.4334          & 0.6876    & 0.6116          & 0.7716    & 0.0000            & 0.6421         & 0.5664   & 0.6361       & 0.7438            & \textbf{0.7798} \\
west\_2001          & 49         & 1198                  & 2                  & 0.1527          & 0.0002    & 0.0886          & 0.1058    & 0.0000            & 0.1715         & 0.3061   & 0.4761       & \textbf{0.5613}   & 0.5035           \\
pomeroy\_2002\_v2   & 42         & 857                   & 5                  & 0.0000                  & 0.4924    & 0.0000                  & 0.0000            & 0.0000            & 0.0514         & 0.3583   & 0.4685       & 0.5175            & \textbf{0.5446} \\
singh\_2002         & 102        & 339                   & 2                  & 0.0259           & 0.0330   & 0.0259           & 0.0330    & 0.0000            & 0.0574         & 0.0330  & 0.3433       & 0.0483           & \textbf{0.8135} \\
tomlins\_v2         & 92         & 1288                  & 4                  & 0.1418          & 0.1245    & 0.0100          & 0.2134    & 0.0000            & 0.1814           & 0.1730   & 0.2775       & 0.1993            & \textbf{0.3985} \\
alizadeh\_2000\_v1  & 42         & 1095                  & 2                  & 0.0127          & 0 .0000           & 0.0000                  & 0.2564    & 0.0000            & 0.0023         & 0.0613   & 0.2714      & 0.0889             & \textbf{0.3716}  \\
armstrong\_2002\_v2 & 72         & 2194                  & 3                  & 0.5123          & 0.5448     & 0.6625          & 0.4584    & 0.0000            & 0.4660         & 0.4992   & 0.6365       & \textbf{0.9186} & 0.8332          \\
bredel\_2005        & 50         & 832                   & 3                  & 0.2000          & 0.3525    & 0.0098          & 0.4760    & 0.0000            & 0.0893         & 0.2315   & 0.4877       & 0.2996            & \textbf{0.5841}\\
\addlinespace
\midrule
Average Rank        &          &                    &                   & 7.2500          & 6.0000  & 7.7500 & 5.3125          & 9.6875    & 5.8750            & 5.7500         & 3.1250          &  3.0000           & \textbf{1.2500}\\
\bottomrule
\end{tabular}}
\label{table:compcancer}
\end{table*}

\subsubsection{Friedman's Rank Test} 

Friedman's rank test \citep{friedman} is employed to discern whether a significant difference exists in the performance of the algorithms applied to our datasets. This assessment unfolds across three stages. In the initial stage, the test encompasses all clustering algorithms under consideration. Moving to the second stage, the analysis omits DP-MoM while incorporating the other algorithms. In the third stage, both MoMPKM and DP-MoM are excluded from the test. The calculated p-values for these three stages are as follows: $1.57 \times 10^{-7}$, $0.0021$, and $0.0599$ respectively. The null hypothesis, which posits no significant variance in clustering accuracy among the tested algorithms, is rejected in the first and second stages. However, it is accepted in the third stage. This outcome underscores that MoMPKM and DP-MoM emerge as the most proficient clustering algorithms at our disposal. Further assessments indicate that MoMPKM is outperformed comprehensively by the novel DP-MoM. 

\subsubsection{Sign Test and Wilcoxon Signed Rank (WSR) Test}

We also perform the \textit{Sign Test} and \textit{Wilcoxon Signed Rank Test} to compare our proposed algorithm individually with every other competing algorithm mentioned in Tables 1 and 2 and check whether DP-MoM performs significantly better than each of the aforementioned state-of-the-art algorithms. It is evident from the $p$-values that the null hypotheses: $H_{0s}:$ The said algorithm is better than our proposed algorithm DP-MoM (\textit{Sign Test}) and $H_{0w}:$ The said algorithm is equivalent to our proposed framework DP-MoM (\textit{Wilcoxon's signed Rank Test}) are rejected in favor of the alternative $H_1:$ DP-MoM performs significantly better than the other state-of-the-art clustering algorithm in question for a test with level of significance $0.01$. In a majority of the cases, our proposed DP-MoM algorithm is the standout performer. However, in the 3 cases where its performance is slightly suboptimal with respect to that of MoMPKM and OWL $k$-means, the results of the statistical tests presented in Table 3, indicate that this drop in performance is not statistically significant at the specified level.

\begin{table}[!htb]
\renewcommand\theadfont{\bfseries}
\centering
\caption{Summary of the Statistical Test Results for level of significance $0.01$}
\vskip 0.7em
\resizebox{0.6\linewidth}{!}{
\begin{tabular}{lccccccccccccc}
\toprule
\multirow[b]{2}{*}{\thead{\textbf{Clustering Algorithm}}}
    &   \multicolumn{2}{c}{\thead{\textbf{Sign Test}}}
        & \multicolumn{2}{c}{\thead{\textbf{WSR Test}}}
             \\
\cmidrule(lr){2-3}\cmidrule(lr){4-5} %\cmidrule(lr){11-13}
% \toprule
% & $\mathbf{n}$ & $\mathbf{d}$ & $\mathbf{K}$ & \thead{KM++} & \thead{DPM} & \thead{KMed} & \thead{PAM} & \thead{K-bMoM} & \\
& \thead{Statistic} & \thead{\textit{\textbf{p}}-value}  & \thead{Statistic} & \thead{\textit{\textbf{p}}-value} 
\\
% \addlinespace
\midrule
\addlinespace
$k$-means ++           & 15  & 0.0002594  & 135 & 0.0000305   \\
Sparse $k$-means          & 15  & 0.0002594  & 135 & 0.0000305          \\
$k$-medians           & 15  & 0.0002594  & 135 & 0.0000305         \\
Partition Around Medoids        & 15  & 0.0002594  & 134 & 0.0000458 \\
Robust Continuous Clustering           & 15  & 0.0002594  & 135 & 0.0000305 \\
DP means       & 15  & 0.0002594  & 133 & 0.0000763 \\
Kb MoM          & 15  & 0.0002594  & 135 & 0.0000305 \\
MoM Power $k$-means & 15  & 0.0002594  & 135 & 0.0000305 \\
OWL $k$-means & 14  & 0.0020900  & 125 & 0.0008392\\
\addlinespace
\bottomrule
\end{tabular}
}
\label{table:statistical tests}
\end{table}

Tables 4 and 5 provide the range of the penalty parameter $\lambda$ that enables us to cluster each dataset more efficiently. The predicted number of clusters are also displayed. Note that the number of clusters have been calculated after assigning the data points in the clusters containing less than $3$ observations, to the nearest cluster containing at least $3$ observations.

\begin{table}[!htb]
\renewcommand\theadfont{\bfseries}
\centering
\caption{Range of optimal $\lambda$ and estimated number of clusters for implementing DP-MoM on UCI datasets}
\vskip 1em
\resizebox{0.5\linewidth}{!}{
\begin{tabular}{lcc}
\toprule
{\thead{\textbf{Dataset}}}
    &   \multicolumn{1}{c}{\thead{\textbf{Range of Optimal $\lambda$}}}
        & \multicolumn{1}{c}{{\textbf{Estimated Clusters}}}\\
% \addlinespace
\midrule
\addlinespace
Iris          & 5.129 - 6.535  & 3 \\
Glass         & 4.207 - 18.841  &  6 \\
WDBC           & 2.5$\times 10^6$ - 6$\times 10^6$  & 2 \\
E. Coli       & 0.3008 - 0.3571  & 8 \\
Wine         &  3.78$\times 10^5$ - 3.93$\times 10^5$ & 2 \\
Thyroid       & 1769 - 2123  & 5 \\
Zoo          & 7.048 - 10.360  & 6 \\
soybean & 18.59 - 25.70  & 4 \\
\addlinespace
\bottomrule
\end{tabular}
}
\label{table:lambda-uci}
\end{table}

\begin{table}[!htb]
\renewcommand\theadfont{\bfseries}
\centering
\caption{Range of optimal $\lambda$ and estimated number of clusters for implementing DP-MoM on Compcancer datasets}
\vskip 1em
\resizebox{0.6\linewidth}{!}{
\begin{tabular}{lcc}
\toprule
{\thead{\textbf{Dataset}}}
    &   \multicolumn{1}{c}{\thead{\textbf{Range of Optimal $\lambda$}}}
        & \multicolumn{1}{c}{\thead{\textbf{Estimated Clusters}}}\\
% \addlinespace
\midrule
\addlinespace
golub\_1999\_v2          & 2.48$\times 10^9$ - 3.05$\times 10^9$ & 3   \\
west\_2001         & 1.77$\times 10^9$  - 3.11$\times 10^9$ &  2         \\
pomeroy\_2002\_v2           & 2.94$\times 10^9$ - 4.70$\times 10^9$ & 4         \\
singh\_2002       & 0.14$\times 10^9$ - 0.17$\times 10^9$ & 2 \\
tomlins\_v2         &  175.6 - 308.0 & 2 \\
alizadeh\_2000\_v1       & 762.8 - 780.0  & 2 \\
armstrong\_2002\_v2          & 5.85$\times 10^9$ - 7.08$\times 10^9$ & 3 \\
bredel\_2005 & 672.4 - 978.8  & 2 \\
\addlinespace
\bottomrule
\end{tabular}
}
\label{table:lambda-compcancer}
\end{table}

\section{Discussion}

In this article, we proposed a new clustering algorithm that integrates two major clustering paradigms, viz., centroid-based and model-based clustering, and is intended to perform well on noisy or outlier-laden data. We utilize the Median-of-Means (MoM) estimator to deal with noise or outliers present in data, and a Bayesian nonparametric modelling ensures that the number of clusters need not be specified. Unlike conventional clustering methods, which tackle only one of these challenges, our proposed algorithm adeptly tackles both at the same time. Following our comprehensive theoretical analysis of error rate bounds, augmented by extensive simulation studies and real-world data analysis, we showcase the superiority of our method against quite a few prominent clustering algorithms. 

However, our proposed technique is not without its shortcomings; the parameter $\lambda$ is notoriously hard to tune without resorting to grid searching using clustering efficiency. On the bright side, our technique works quite well even in the high-dimensional setting, and it might be fruitful to explore convergence results in this regime to further enhance its applicability and efficacy in the future.
\bibliographystyle{apalike}
\bibliography{main_v2}

%%%%%%%%%%%%%%%%%%%%%%%%%%%%%%%%%%%%%%%%%%%%%%%%%%%%%%%%%%%%
\appendix

\newpage

\begin{center}
    \Large\textbf{Supplementary Material}    
\end{center}

\vspace{0.2cm}

\section{Proofs of Results in Section \ref{sec:theory-new}}
\subsection{Proof of Theorem \ref{thm-4-MoM}}
\begin{proof}
    We represent the empirical distribution of $\{\bX_i\}_{i \in B_\ell}$ by $P_{B_{\ell}}$. Fix $\varepsilon>0$. We will first establish a concentration inequality on $\sup_{\bTheta \in \mathscr{G}} |\text{MoM}_L^n (f_{\bTheta}) - P f_{\bTheta} |$ where $\mathscr{G}$ denotes the collection of all $\bTheta$ such that $\|\btheta\|\le M$ for all $\btheta\in \bTheta$. To that end, we bound the probabilities of the events $\{\sup_{\bTheta \in \mathscr{G}}(\text{MoM}_L^n (f_{\bTheta}) - P f_{\bTheta}) >\varepsilon\}$ and $\{\sup_{\bTheta \in \mathscr{G}}  ( P f_{\bTheta} - \text{MoM}_L^n (f_{\bTheta})) > \varepsilon\}$ individually. Note that 
\begin{equation}
\sup_{\bTheta \in \mathscr{G}}\sum_{\ell = 1}^L \one\left\{(P-P_{B_\ell})f_{\bTheta} > \varepsilon\right\} > \frac{L}{2} \implies \sup_{\bTheta \in \mathscr{G}} (P f_{\bTheta} - \text{MoM}_L^n (f_{\bTheta})) > \varepsilon.
\end{equation}
To see this, suppose on the contrary that \[\sup_{\bTheta \in \mathscr{G}}  (P f_{\bTheta} - \text{MoM}_L^n (f_{\bTheta})) \le \varepsilon\] but \[\sup_{\bTheta \in \mathscr{G}}\sum_{\ell = 1}^L \one\left\{(P-P_{B_\ell})f_{\bTheta} > \varepsilon\right\} > \frac{L}{2}\]. 
Then for all $\bm{\Theta}\in \mathscr{G}$, we must have \[(P f_{\bTheta} - \text{MoM}_L^n (f_{\bTheta})) \le \varepsilon\] which implies that for all $\bm{\Theta}\in \mathscr{G}$ \[\sum_{\ell = 1}^L \one\left\{(P-P_{B_\ell})f_{\bTheta} \le \varepsilon\right\} \ge \frac{L}{2} \implies \sum_{\ell = 1}^L \one\left\{(P-P_{B_\ell})f_{\bTheta} > \varepsilon\right\} \le \frac{L}{2}\] which in turn implies that \[\sup_{\bTheta \in \mathscr{G}}\sum_{\ell = 1}^L \one\left\{(P-P_{B_\ell})f_{\bTheta} > \varepsilon\right\} \le \frac{L}{2}\] which is a contradiction. Let $\varphi(t) = (t-1) \one\{1 \le t \le 2\} + \one\{t >2\}$ where $\one\{\cdot\}$ is the indicator function. Evidently,
\begin{equation}
    \label{eq6}
    \one\{t \ge 2\} \le \varphi(t) \le \one\{t \ge 1\}.
\end{equation} 
We observe that
\begingroup
\allowdisplaybreaks
\begin{align}
    \sup_{\bTheta \in \mathscr{G}}\sum_{\ell = 1}^L \one\left\{(P-P_{B_\ell})f_{\bTheta} > \varepsilon\right\}
    % \le & \sup_{\bTheta \in \mathscr{G}}\sum_{\ell \in \cL} \one\left\{(P-P_{B_\ell})f_{\bTheta} > \varepsilon\right\} + |\cO|\nonumber\\
    % \le & \sup_{\bTheta \in \mathscr{G}}\sum_{\ell \in \cL}  \varphi\left(\frac{2(P-P_{B_\ell})f_{\bTheta}}{\varepsilon}\right) + |\cO|\nonumber\\
    & \le \sup_{\bTheta \in \mathscr{G}}\sum_{\ell \in \cL}  \E \varphi\left(\frac{2(P-P_{B_\ell})f_{\bTheta}}{\varepsilon}\right)  + |\cO|\nonumber\\
    & ~+\sup_{\bTheta \in \mathscr{G}}\sum_{\ell \in \cL}  \bigg[ \varphi\left(\frac{2(P-P_{B_\ell})f_{\bTheta}}{\varepsilon}\right)   - \E \varphi\left(\frac{2(P-P_{B_\ell})f_{\bTheta}}{\varepsilon}\right)\bigg]. \label{eq01}
\end{align}
\endgroup
Towards bounding $\sup_{\bTheta \in \mathscr{G}}\sum_{\ell = 1}^L \one\left\{(P-P_{B_\ell})f_{\bTheta} > \varepsilon\right\}$, we first bound $\E \varphi\left(\frac{2(P-P_{B_\ell})f_{\bTheta}}{\varepsilon}\right)$. Note that
\begingroup
\allowdisplaybreaks
% \begin{align}
% \small 
%   \E \varphi\left(\frac{2(P-P_{B_\ell})f_{\bTheta}}{\varepsilon}\right) 
%   \le  \E \left[ \one\left\{\frac{2(P-P_{B_\ell})f_{\bTheta}}{\varepsilon} > 1 \right\}\right] \nonumber  = & \sP \left[ (P-P_{B_\ell})f_{\bTheta}>\frac{\varepsilon}{2} \right] \nonumber \\ 
%   \le & \exp\left\{-\frac{b \varepsilon^2}{32 M^4 K^2 p^2}\right\}
% \end{align} 
\begin{align*}
    \E \varphi\left(\frac{2(P-P_{B_\ell})f_{\bTheta}}{\varepsilon}\right) \le  %\E \left[ \one\left\{\frac{2(P-P_{B_\ell})f_{\bTheta}}{\varepsilon} > 1 \right\}\right] = 
    \sP \left[ (P-P_{B_\ell})f_{\bTheta}>\frac{\varepsilon}{2} \right]
    &\le \frac{4}{\varepsilon^2}\text{Var}P_{B_{\ell}} f_{\bTheta}\\
    &=\frac{4L}{|\mathcal{I}|\varepsilon^2}\E\min_{\btheta\in \bTheta}(\|\btheta\|^2-2\btheta^\top \bX)^2\\
    &\le \frac{4L}{|\mathcal{I}|\varepsilon^2}M^2(M+2\gamma)^2.
\end{align*}

The last inequality follows by Chebyshev's inequality after observing that $\displaystyle \mathbb{E} P_{B_{\ell}}f_{\bm{\Theta}}=Pf_{\bm{\Theta}}$.
\endgroup
 We now turn to bounding the term \[ \sup_{\bTheta \in \mathscr{G}}\sum_{\ell \in \cL}  \bigg[ \varphi\left(\frac{2(P-P_{B_\ell})f_{\bTheta}}{\varepsilon}\right)   - \E \varphi\left(\frac{2(P-P_{B_\ell})f_{\bTheta}}{\varepsilon}\right)\bigg]. \] Appealing to Theorem 26.5 of \cite{shalev-shwartz_ben-david_2014}, for all $ \bTheta \in \mathscr{G}$, we have
\begin{align}
  &\frac{1}{L}\sum_{\ell \in \cL}   \varphi\left(\frac{2(P-P_{B_\ell})f_{\bTheta}}{\varepsilon}\right)- \E \left[\frac{1}{L}\sum_{\ell \in \cL}  \varphi\left(\frac{2(P-P_{B_\ell})f_{\bTheta}}{\varepsilon}\right) \right]\nonumber \\
  &\le 2\E\left[\sup_{\bTheta \in \mathscr{G}}\frac{1}{L}\sum_{\ell \in \cL}\sigma_\ell  \varphi\left(\frac{2(P-P_{B_\ell})f_{\bTheta}}{\varepsilon}\right) \right] + \delta.  \label{eq5}
\end{align}
with probability at least $1-e^{-2L\delta^2}$,
where $\{\sigma_\ell\}_{\ell \in \mathcal{L}}$ are i.i.d. Rademacher random variables independent of the sample $\mathcal{X}$. Consider i.i.d. Rademacher random variables $\{\xi_i\}_{i=1}^n$ independent of  $\{\sigma_\ell\}_{\ell \in \mathcal{L}}$. As a consequence of equation \eqref{eq5}, we get,  
\begingroup
\allowdisplaybreaks
\begin{align}
     & \frac{1}{L}\sup_{\bTheta \in \mathscr{G}}\sum_{\ell \in \cL}  \bigg[ \varphi\left(\frac{2(P-P_{B_\ell})f_{\bTheta}}{\varepsilon}\right)  - \E \varphi\left(\frac{2(P-P_{B_\ell})f_{\bTheta}}{\varepsilon}\right)\bigg] \nonumber\\
     & \le \frac{4}{L \varepsilon}\E\left[\sup_{\bTheta \in \mathscr{G}}\sum_{\ell \in \cL}\sigma_\ell  (P-P_{B_\ell})f_{\bTheta} \right]+ \delta. \label{eq7} 
     \end{align}
     Equation \eqref{eq7} follows from the fact that $\varphi(\cdot)$ is 1-Lipschitz and Lemma 26.9 of \cite{shalev-shwartz_ben-david_2014}. We now consider random variables $\mathcal{X}^\prime=\{\bX_1^\prime, \dots, \bX_n^\prime\}$, which are i.i.d. and follow the law $P$. Equation \eqref{eq7} further yields
     \begin{align}
     = & \frac{4}{L \varepsilon}\E\left[\sup_{\bTheta \in \mathscr{G}}\sum_{\ell \in \cL}\sigma_\ell  \E_{\mathcal{X}^\prime}\left((P^\prime_{B_\ell}-P_{B_\ell})f_{\bTheta}\right) \right]+ \delta \nonumber \\
     \le & \frac{4}{L \varepsilon}\E\left[\sup_{\bTheta \in \mathscr{G}}\sum_{\ell \in \cL}\sigma_\ell  (P^\prime_{B_\ell}-P_{B_\ell})f_{\bTheta} \right]+ \delta \nonumber \\
     & \text{This inequality follows by employing Jensen's inequality as supremum is a convex function,}\nonumber\\ & \text{and the expectation in the second step is taken with respect to both the original sample $\mathcal{X}$}\nonumber\\ & \text{as well as the ghost sample $\mathcal{X}'$ and the Rademacher random variables $\{\sigma_{\ell}\}_{\ell\ge 1}$.}\nonumber\\
     = & \frac{4}{L \varepsilon}\E\left[\sup_{\bTheta \in \mathscr{G}}\sum_{\ell \in \cL}\sigma_\ell  \frac{1}{b}\sum_{i \in B_\ell}(f_{\bTheta}(\bX_i^\prime)-f_{\bTheta}(\bX_i)) \right]+ \delta \nonumber \\
     = & \frac{4}{b L \varepsilon}\E\left[\sup_{\bTheta \in \mathscr{G}}\sum_{\ell \in \cL}\sigma_\ell  \sum_{i \in B_\ell} \xi_i(f_{\bTheta}(\bX_i^\prime)-f_{\bTheta}(\bX_i)) \right]+ \delta \label{eq8} \\
     = & \frac{4}{|\mathcal{I}| \varepsilon}\E\left[\sup_{\bTheta \in \mathscr{G}}\sum_{\ell \in \cL}  \sum_{i \in B_\ell} \sigma_\ell \xi_i(f_{\bTheta}(\bX_i^\prime)-f_{\bTheta}(\bX_i)) \right]+ \delta \nonumber \\
     = & \frac{4}{|\mathcal{I}| \varepsilon}\E\left[\sup_{\bTheta \in \mathscr{G}} \sum_{i \in \J} \gamma_i (f_{\bTheta}(\bX_i^\prime)-f_{\bTheta}(\bX_i)) \right]+ \delta \label{eq9} \\
     \le &  \frac{4}{|\mathcal{I}| \varepsilon}\E\left[\sup_{\bTheta \in \mathscr{G}} \sum_{i \in \J} \gamma_i f_{\bTheta}(\bX_i^\prime)+ \sup_{\bTheta \in \mathscr{G}} \sum_{i \in \J} \gamma_i f_{\bTheta}(\bX_i) \right]+ \delta\\
     \le & \frac{4}{|\mathcal{I}| \varepsilon}\E\left[\sup_{\bTheta \in \mathscr{G}} \sum_{i \in \J} \gamma_i f_{\bTheta}(\bX_i^\prime)\right]+\E\left[ \sup_{\bTheta \in \mathscr{G}} \sum_{i \in \J} \gamma_i f_{\bTheta}(\bX_i) \right]+ \delta\\
     = & \frac{8}{ \varepsilon}\E\left[\sup_{\bTheta \in \mathscr{G}} \frac{1}{|\mathcal{I}|}\sum_{i \in \J} \gamma_i f_{\bTheta}(\bX_i) \right]+ \delta. \label{eq10}
\end{align}
\endgroup

Here $\J$ is the set of observations in the partitions not containing an outlier. Equation \eqref{eq8} follows from the fact that  $(f_{\bTheta}(\bX_i^\prime)-f_{\bTheta}(\bX_i)) \overset{d}{=} \xi_i(f_{\bTheta}(\bX_i^\prime)-f_{\bTheta}(\bX_i))$ as $\{\xi_i\}_{i\ge 1}$ are Rademacher random variables independent of the sample $\mathcal{X}$. In equation \eqref{eq9}, $\{\gamma_i\}_{i \in \mathcal{J}}$ are independent Rademacher random variables since the product of two independent Rademacher random varibles is also a Rademacher random variable. 

We compute the Rademacher complexity in Eq \eqref{eq10} as follows:
Note that $|\Psi(\bx)-\Psi(\by)|\le \|\bx-\by\|_2$. To see this, observe that $|\Psi(\bx)-\Psi(\by)|\le |\min_{1\le j\le k}x_j-\min_{1\le i\le k}y_i|.$ Now, without loss of generality, let $\min_{1\le j\le k}x_j\ge\min_{1\le i\le k}y_i$. Then $|\Psi(\bx)-\Psi(\by)|\le x_{j^*}-y_{j^*}\le \|\bx-\by\|_2$ where $j^*=\argmin_{1\le i\le k}y_i$. Let $g_{\bTheta}(\bx)=(-2\langle\bx,\btheta_1\rangle+\|\btheta_1\|^2,\ldots,-2\langle\bx,\btheta_k\rangle+\|\btheta_k\|^2)$, we have \[|f_{\bTheta}(\bX_i)-f_{\bTheta}(\bX_i)|\le \|g_{\bTheta}(\bX_i)-g_{\bTheta}(\bX_i)\|_2.\]
Further Maurer's vector contraction inequality yields \[\E\left[\sup_{\bTheta \in \mathscr{G}} \frac{1}{|\mathcal{I}|}\sum_{i \in \J} \gamma_i f_{\bTheta}(\bX_i) \right]\le \frac{\sqrt{2}}{|\mathcal{I}|}\left(2\E \sup_{\bTheta\in\mathscr{G}}\sum_{j=1}^{k}\sum_{i=1}^{|\mathcal{I}|}\varepsilon_{i,j}\langle \bX_i,\btheta_j\rangle+\E \sup_{\bTheta\in\mathscr{G}}\sum_{j=1}^{k}\sum_{i=1}^{|\mathcal{I}|}\varepsilon_{i,j}\|\btheta_j\|^2\right).\]
Moreover, by Khintchine's Inequality, we have \begin{align*}
    \E \sup_{\bTheta\in\mathscr{G}}\sum_{j=1}^{k}\sum_{i=1}^{|\mathcal{I}|}\varepsilon_{i,j}\langle \bX_i,\btheta_j\rangle &\le \sum_{j=1}^k \E\sup_{\bTheta\in\mathscr{G}}\left\langle\sum_{i=1}^{|\mathcal{I}|}\varepsilon_{i,j}\bX_i,\btheta_j\right\rangle\\
    &\le \sum_{j=1}^k \E\sup_{\bTheta\in\mathscr{G}}\left\|\sum_{i=1}^{|\mathcal{I}|}\varepsilon_{i,j}\bX_i\right\|\|\btheta_j\|\\
    &\le kM\max_{j\le k}\E \left\|\sum_{i=1}^{|\mathcal{I}|}\varepsilon_{i,j}\bX_i\right\|\\
    &\le kM\max_{j\le k}\sqrt{\E\left\|\sum_{i=1}^{|\mathcal{I}|}\varepsilon_{i,j}\bX_i\right\|^2}\\
    &\le kM\sqrt{\sum_{i=1}^{|\mathcal{I}|}\E\|\bX_i\|^2}\le kM\gamma\sqrt{|\mathcal{I}|}
\end{align*}
and \[\E \sup_{\bTheta\in\mathscr{G}}\sum_{j=1}^{k}\sum_{i=1}^{|\mathcal{I}|}\varepsilon_{i,j}\|\btheta_j\|^2\le \sum_{j=1}^k \E\sup_{\bTheta\in \mathscr{G}}\left\|\sum_{i=1}^{|\mathcal{I}|}\varepsilon_{i,j}\right|\|\theta_j\|^2\le kM^2\sqrt{|\mathcal{I}|}.\]

Thus, we have \[\E\left[\sup_{\bTheta \in \mathscr{G}} \frac{1}{|\mathcal{I}|}\sum_{i \in \J} \gamma_i f_{\bTheta}(\bX_i) \right]\le \frac{\sqrt{2}kM(M+2\gamma)}{\sqrt{|\mathcal{I}|}}.\]

From the above computations, we conclude that,  with probability of at least $1-e^{-2L \delta^2}$, 
\begin{equation}
    \sup_{\bTheta \in \mathscr{G}}\sum_{\ell = 1}^L \one\left\{(P-P_{B_\ell})f_{\bTheta} > \varepsilon\right\} \le  L \left( \frac{4}{|\mathcal{I}|\varepsilon^2}M^2(M+2\gamma)^2+ \frac{|\cO|}{L} + \frac{8\sqrt{2}kM(M+2\gamma)}{\varepsilon\sqrt{|\mathcal{I}|}} + \delta\right). \label{eq12}
\end{equation}
We choose $\delta = \frac{2}{4+\eta} - \frac{|\cO|}{L}$, and $$\varepsilon = \max\left\{\frac{4\sqrt{(\eta+4)}}{\sqrt{\eta }\sqrt{n}}M(M+2\gamma), \frac{32(\eta+4)}{\eta\sqrt{n}}M(M+2\gamma)\right\}.$$ This makes the right hand side of \eqref{eq12} strictly smaller than $\frac{L}{2}$.
Thus, we have shown that 
\begin{align*}
     \sP\left( \sup_{\bTheta \in \mathscr{G}} (Pf_{\bTheta} - \text{MoM}^n_L (f_{\bTheta}))> \varepsilon \right) \le e^{-2L \delta^2}.
 \end{align*}
% \begin{align*}
%     &P\left( \sup_{\bTheta \in \mathscr{G}} (Pf_{\bTheta} - \text{MoM}^n_L (f_{\bTheta}))> \varepsilon \right) \\
%     & \le P\left(\sup_{\bTheta \in \mathscr{G}}\sum_{\ell = 1}^L \one\left\{(P-P_{B_\ell})f_{\bTheta} > \varepsilon\right\} \ge L/2\right) \le e^{-2L \delta^2}.
% \end{align*}
By a similar argument, it follows that,
\begin{align*}
    \sP\left( \sup_{\bTheta \in \mathscr{G}} (\text{MoM}^n_L (f_{\bTheta}) -Pf_{\bTheta} ) > \varepsilon \right) \le e^{-2L \delta^2}.
\end{align*}
From the two aforementioned inequalities, we obtain, 
\[\sP\left( \sup_{\bTheta \in \mathscr{G}} |\text{MoM}^n_L (f_{\bTheta}) -Pf_{\bTheta} | > \varepsilon \right) \le 2e^{-2L \delta^2}.\]
i.e., with at least probability $1-2e^{-2L \delta^2}$,
\begin{align*}
    & \sup_{\bTheta \in \mathscr{G}} |\text{MoM}^n_L (f_{\bTheta}) -Pf_{\bTheta} |\\ &\le \max\left\{\frac{4\sqrt{(\eta+4)}}{\sqrt{\eta}\sqrt{n}}M(M+2\gamma), \frac{32(\eta+4)}{\eta\sqrt{n}}M(M+2\gamma)\right\}\lesssim \frac{M(M+2\gamma)}{\sqrt{n}}.
\end{align*}
\end{proof}

\subsection{Proof of Corollary \ref{cor-mom-1}}
\begin{proof}
    Note that the following holds with probability at least $1-2e^{-2L\delta^2}$:
    \begin{align*}
        &\left|P f_{\widehat{\bTheta}_n^{\text{MoM}}}-P f_{\bTheta^*}\right|\\
        &= P f_{\widehat{\bTheta}_n^{\text{MoM}}}-P f_{\bTheta^*}\\
        &= P f_{\widehat{\bTheta}_n^{\text{MoM}}}-\text{MoM}_L^n\left(f_{\widehat{\bTheta}_n^{\text{MoM}}}\right)+\text{MoM}_L^n\left(f_{\widehat{\bTheta}_n^{\text{MoM}}}\right)-\text{MoM}_L^n(f_{\bTheta^*})+\text{MoM}_L^n(f_{\bTheta^*})-P f_{\bTheta^*}\\
        &\le P f_{\widehat{\bTheta}_n^{\text{MoM}}}-\text{MoM}_L^n\left(f_{\widehat{\bTheta}_n^{\text{MoM}}}\right)+\text{MoM}_L^n(f_{\bTheta^*})-P f_{\bTheta^*}\\
        &\le 2\sup_{\bTheta\in \mathscr{G}}\left|\text{MoM}_L^n(f_{\bTheta})-P f_{\bTheta}\right|\lesssim M(M+2\gamma)n^{-1/2}
    \end{align*}
    Note that the penultimate inequality follows by the definition of $\widehat{\bTheta}_n^{\text{MoM}}$.
\end{proof}

\subsection{Proof of Corollary \ref{cor-mom-2}}
\begin{proof}
    Since $\delta =2/(4+\eta)-|\mathcal{O}|/L>\eta/(4+\eta)(2+\eta)$ which implies that $e^{-2L\delta^2}=o(1)$ as $n\to \infty$. Thus, \[\mathbb{P}\left(\left|P f_{\widehat{\bTheta}_n^{\text{MoM}}}-P f_{\bTheta^*}\right|=\mathcal{O}(n^{-1/2})\right)\ge 1-o(1)\] which implies that $\left|P f_{\widehat{\bTheta}_n^{\text{MoM}}}-P f_{\bTheta^*}\right|=\mathcal{O}_P(n^{-1/2})$. 
    
    Further, as $n\to \infty$, we must have $\left|P f_{\widehat{\bTheta}_n^{\text{MoM}}}-P f_{\bTheta^*}\right|=o_P(n^{-1/2})$ which in turn implies that $P f_{\widehat{\bTheta}_n^{\text{MoM}}}\overset{p}{\longrightarrow}P f_{\bTheta^*}$. 
    
    Given this, for every $\delta,\varepsilon>0$, $\exists N\in \mathbb{N}$ such that $\forall$ $n\ge N$, we have $\mathbb{P}\left(P f_{\widehat{\bTheta}_n^{\text{MoM}}}-P f_{\bTheta^*}> \varepsilon\right)< \delta$ which means that for any $\eta>0$, $\exists$ $\varepsilon>0$ such that  $\mathbb{P}\left(\mathcal{D}\left(\bTheta_n^{\text{MoM}},\bTheta^*\right)> \eta\right)<\delta$ and hence, we may conclude that $\bTheta_n^{\text{MoM}}\overset{p}{\longrightarrow}\bTheta^*$.
\end{proof}

\end{document}